# Deep learning-based object detection of offshore platforms on Sentinel-1 Imagery and the impact of synthetic training data

Robin Spanier*[a], Thorsten Hoeser[a], and Claudia Kuenzer[a,b]

[a]*German Remote Sensing Data Center, Earth Observation Center, EOC of the German Aerospace Center, DLR, Wessling, Germany*
[b]*Institute for Geography and Geology, Department of Remote Sensing, University of Würzburg, Würzburg, Germany*

*Corresponding author: robin.spanierdlr.de*

**Abtract**

**The recent and ongoing expansion of marine infrastructure, including offshore wind farms, oil and gas platforms, artificial islands, and aquaculture facilities, highlights the need for effective monitoring systems. The development of robust models for offshore infrastructure detection relies on comprehensive, balanced datasets, but falls short when samples are scarce, particularly for underrepresented object classes, shapes, and sizes. By training deep learning-based YOLOv10 object detection models with a combination of synthetic and real Sentinel-1 satellite imagery acquired in the fourth quarter of 2023 from four regions (Caspian Sea, South China Sea, Gulf of Guinea, and Coast of Brazil), this study investigates the use of synthetic training data to enhance model performance. We evaluated this approach by applying the model to detect offshore platforms in three unseen regions (Gulf of Mexico, North Sea, Persian Gulf) and thereby assess geographic transferability. This region-holdout evaluation demonstrated that the model generalises beyond the training areas. In total, 3,529 offshore platforms were detected, including 411 in the North Sea, 1,519 in the Gulf of Mexico, and 1,593 in the Persian Gulf. The model achieved an F1 score of 0.85, which improved to 0.90 upon incorporating synthetic data. We analysed how synthetic data enhances the representation of unbalanced classes and overall model performance, taking a first step toward globally transferable detection of offshore infrastructure. This study underscores the importance of balanced datasets and highlights synthetic data generation as an effective strategy to address common challenges in remote sensing, demonstrating the potential of deep learning for scalable, global offshore infrastructure monitoring.**

## 1. Introduction

In the past few decades, the expansion of maritime persistent infrastructure has accelerated, particularly in relation to energy infrastructure such as oil and gas platforms and wind farms [1, 2]. Among these, offshore platforms play a central role in supporting a wide range of marine operations. Platforms serve as a base for offshore activities, provide work and living facilities, and are essential for the exploitation of marine resources, research, production, rescue operations, maritime logistics, storage, and transport [3]. They occur in various structural configurations adapted to water depth and environmental conditions, designed to operate for over 25 years in harsh environments (Figure 1) [4]. This maritime infrastructure contributes significantly to economic development and drive social and technological progress in regions that have previously been poor in resources. In 2020, offshore oil and gas extraction reached a peak at USD 988 billion in gross output value [5]. Stakeholders such as governments, regulatory authorities, and operators seek to harmonise international installation and decommissioning practices of this infrastructure [5, 6]. The aim is to effectively manage the dynamic development of the maritime space while taking into account the associated ecological, economic, and social interests, for example in the areas of ecosystem services, fisheries, logistics, shipping, the recreation industry, and nature conservation [7–10]. Reliable data on offshore infrastructure is essential for maritime domain awareness and facilitates environmental management, decision making, and policy to address challenges across their life cycle [5, 6, 11]. Some installations transmit positional data via AIS to enhance monitoring [12], and regulations such as the EU's Environmental Impact Assessment (EIA) mandate reporting of new facilities [13]. However, such regulations are regionally limited and inconsistently enforced, and existing data are often incomplete or restricted due to confidentiality and security concerns [14–17]. Given these challenges, the development of reliable, scalable methods for accurately and quickly detecting the location and characteristics, such as size, type, or design, of offshore infrastructure is a key concern.

Over the past 50 years, tens of millions of satellite images have been collected and archived, opening up new possibilities for environmental and infrastructure monitoring. Advances in this field provide an overview of the condition and dynamics of large-scale Earth surfaces [18–20]. Satellite remote sensing offers regular and effective monitoring without the need for direct access, particularly for regions that are difficult to access, such as offshore areas [21, 22]. With a short repeat cycle, global coverage, and the continuous availability of long-term measurements, a wide range of data is collected by various sensors and made publicly available in a constantly growing, updated, and nearly global database. In addition, earth observation (EO) has benefited significantly from advances in cloud computing and data science. Platforms such as Google Earth Engine (GEE) make it possible to efficiently analyse large amounts of remote sensing data without having to store or download it locally [23, 24].

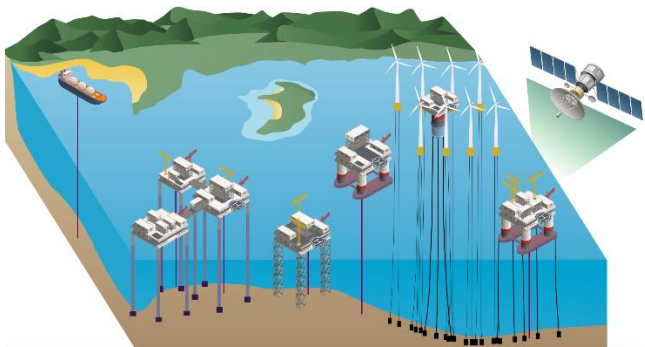

**Figure 1.** Schematic overview of offshore energy infrastructure, in particular, different platform types. Fixed platforms are anchored to the seabed using piles or jackets and are common in shallow waters, while semi-permanent structures are partially mobile and used in deeper settings. Figure adjusted from [25].

Studies have been conducted on the detection of permanent man-made marine infrastructure using remote sensing data to investigate its conditions and developments. While, especially in the early phases of research, the high thermal and optical contrast of gas flares was used to identify platforms on infrared images [26], current studies mostly utilise synthetic aperture radar (SAR) data. SAR is recognised as a highly effective remote sensing technology for maritime environment monitoring, due to its ability to operate under all weather conditions and strong sensitivity to metallic sea-surface targets such as ships, platforms, and wind turbines. Cheng et al. (2013) demonstrated the use of multitemporal SAR data, combined with a two-parameter constant false alarm rate (CFAR) detector and a triangular invariant rule, to detect offshore platforms in the South China Sea [27]. Marino et al. (2017) analysed the multipolarisation backscattering behaviour of offshore platforms using dual-polarisation X-band SAR imagery [28]. In 2019, Wong et al. developed a geoprocessing algorithm on GEE based on multitemporal SAR median composites to detect oil platforms in the Gulf of Mexico and offshore wind turbines in the exclusive economic zones (EEZ) of the United Kingdom and China [15]. Liu et al. (2019) detected global offshore oil and gas platforms using order statistical filtering with thresholding and Landsat-8 time-series [1]. Zhang et al. (2019) proposed an extraction method from single SAR images, implementing a two-step model that first applies a two-parameter CFAR algorithm to identify potential targets and subsequently uses the Hough transform to eliminate linear ship targets [29]. Xu et al. (2020) processed multispectral images from Sentinel-2 (S2) and Landsat to extract marine infrastructure in the North Sea using order statistical filtering in combination with predefined thresholds [30]. Zhang et al. (2021) extracted offshore wind turbines globally on Sentinel-1 (S1) imagery by applying a morphological approach in combination with multiple thresholds to remove false alarms [31]. In 2022, Hoeser and Kuenzer developed DeepOWT, featuring a cascade of two convolutional neural networks (CNNs) that first detect wind farms and subsequently their turbines on S1 imagery. They derived a global offshore wind turbine dataset and spatiotemporal dynamics of the offshore wind energy sector by deriving the installed capacity [18, 32]. Paolo et al. (2024) utilised S1 SAR data alongside S2 RGB and near-infrared bands to detect ships, wind turbines, and offshore platforms globally [33]. Ma et al. (2024) proposed a method for geolocating and measuring offshore infrastructure by fusing S1 SAR data with S2 multispectral imagery [8]. Qui et al. (2024) integrated Glimmer Imager Pan and RGB bands to detect active platforms in the South China Sea [34]. In 2025, Zhang et al. introduced OPDNet, featuring a two-branch pseudo-Siamese architecture that leverages paired but unsynchronised S1 radar and S2 optical images, showcasing their method by extracting offshore platforms in Shandong Province, China [3].

Most offshore platform detection studies focus on specific regions and use multitemporal approaches to exclude moving targets. Since 2012, roughly 20 % of studies have applied deep learning, though traditional pixel-based methods remain common. However, increasing data availability and computing power are driving a shift toward machine learning and deep learning [25].

The motivation for this study is based on the importance of an optimised training strategy for deep learning-based object detection in EO data. The study evaluates the effects of these optimisations using established YOLO base models, which are known for their efficiency and high detection performance. The central contributions of this work can be divided into four main aspects:

- The incorporation of synthetic data which significantly improves detection performance, especially for underrepresented targets.
- A region holdout strategy enabling model generalisation and geographical transferability of the models beyond the training areas.
- The introduction of a fully trainable offshore platform detection framework based on the GEE and the Google Cloud Platform (GCP), which delivers robust results for hotspot regions of maritime infrastructure.
- A scalable approach that can be applied to the entire S1 archive, enabling global detection and spatiotemporal analysis.

## 2. Materials and methods

### 2.1. Study regions

The study focused on seven regions known for their abundance of offshore energy infrastructure, particularly oil, gas, and wind: the North Sea (NS), Persian Gulf (PG), Gulf of Mexico (GoM), South China Sea (SCS), Caspian Sea (CS), Gulf of Guinea (GoG), and the Coast of Brazil (CoB) (Figure 2). The latter four regions were used to train the object detection (OD) model, while NS, PG, and GoM served for inference. This region-holdout setup enabled evaluation of model generalisation and geographic transferability beyond the training areas. The boundaries of the regions of interest (ROI) followed the International Hydrographic Organisation definitions [35].

### 2.2. Image acquisition and preprocessing

SAR imagery from the European Space Agency (ESA) Copernicus S1 mission [36] was used due to its proven reliability for detecting metallic sea-surface targets, independence of weather, global coverage, high spatial and temporal resolution, and free access [25, 37]. The S1 constellation comprises two C-band SAR satellites. S1A (launched 2014) and S1B (2016), in sun-synchronous orbits with a 180° phase offset, providing a 6-day combined revisit. Due to the failure of S1B in 2021 and the selected time of interest (TOI), the fourth quarter of 2023 (2023Q4), this study used S1A only. Acquisition frequency varies regionally with orbital overlap and mission priorities, resulting in denser coverage over Europe [36] (Figure 2).

We used the Ground Range Detected (GRD) Level-1 product provided by GEE [38], which processes the backscatter coefficient ($\sigma°$) in decibels (dB). Each scene was preprocessed in the following steps: orbit data correction, removal of edge artefacts and thermal noise, radiometric calibration, and terrain correction. For our analysis, we used VH polarisation data in Interferometric Wide (IW) Swath mode, with a spatial resolution of approximately 10 m. The S1 archive was queried for all scenes acquired during the TOI 2023Q4, and for the regions of interest (ROI), the SCS, CS, GoG, and CoB (for model training) and the NS, GoM, and PG (for testing). To enable efficient data handling and export on GEE, we applied a grid over the ROIs (Figure 2). The grid was based on the tiling system of S2 that subdivides Earth into a predefined set of tiles (UTM/WGS84), using a 100 km step. However, each tile has a surface area of 110 x 110 km to provide a large overlap with the neighbouring tiles [39].

These image tiles were preprocessed, using the GEE API, to create deep-learning-ready inputs and to remove transient or mobile objects. First, all image tiles were stacked and reduced to a median composite.

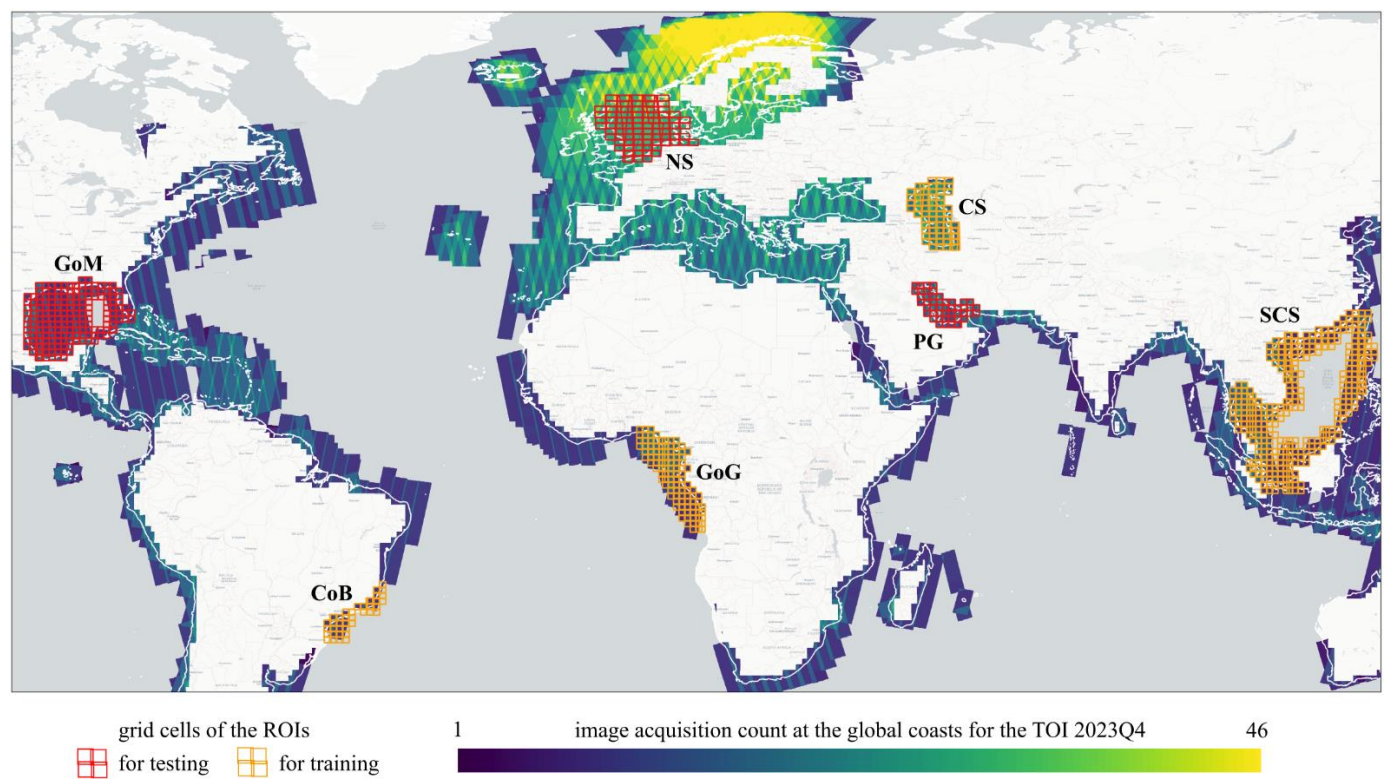

**Figure 2.** Overview of the seven study regions: South China Sea (SCS), the Caspian Sea (CS), the Gulf of Guinea (GoG), and the coast of Brazil (CoB) used for training (orange) and North Sea (NS), the Gulf of Mexico (GoM), and the Persian Gulf (PG) used for testing (red). In addition, the map displays the spatial distribution and image acquisition count of Sentinel-1 SAR scenes that overlap with the global coasts in 2023Q4.

This temporal aggregation effectively removed moving or temporary objects and thus reduced potential false positives even before model inference [15]. A total of 928 median-composite tiles across all study sites were generated and exported (Table 1).

**Table 1.** Satellite imagery downloaded for each study region and 2023Q4.

| Regions of interest | approx. area [km$^2$] | S1 scenes | median comp. tiles |
|---|---|---|---|
| North Sea | 570,000 | 852 | 94 |
| Persian Gulf | 250,000 | 291 | 51 |
| Gulf of Mexico | 1,340,000 | 554 | 199 |
| South China Sea | 2,050,000 | 946 | 363 |
| Caspian Sea | 400,000 | 457 | 69 |
| Gulf of Guinea | 820,000 | 354 | 109 |
| Coast of Brazil | 150,000 | 57 | 44 |
| **Sum ∑** | | 3,511 | 928 |

To minimise data volume, the median composites were converted from 16-bit floating point (backscatter amplitude in dB) to 8-bit integers by clipping and linearly mapping the backscatter signal range of -40 dB to 0 dB onto values 0 to 255. The resulting tiles were then further tiled into 640x640 pixel chips with a 20% overlap, input sizes compatible with the DL model, and to prevent boundary artefacts.

On SAR images, offshore platforms appear as bright backscatter clusters in front of the darker sea with a low backscatter coefficient (Figure 3). The backscatter signal of a platform is larger than its actual size. For example, the signal of a 120 x 70 m platform can be approximately 290 x 230 m on a radar image. The signal is caused by several backscatter effects, including direct reflections from the highest structures of the platform (layover), double reflections between the platform's vertical structures and the sea surface, and triple reflections (or even higher order) between the platform and the surrounding sea surface [28, 40]. Figure 3 illustrates the complexity of platform types and their backscatter signals on S1 SAR images. As mentioned above, they differ in their functionality, shape, and size, and whether they are single platforms or multiple interconnected platforms, as well as in their resulting backscatter signal. We have assigned a dedicated class for wind turbines as off-targets in our dataset. Detecting them is an important step, as their backscatter signals are similar in strength to those of platforms (Figure 3h) and can lead to misclassifications. Due to their distinct radar signature, their detection and separation from platforms is possible [15, 30–32].

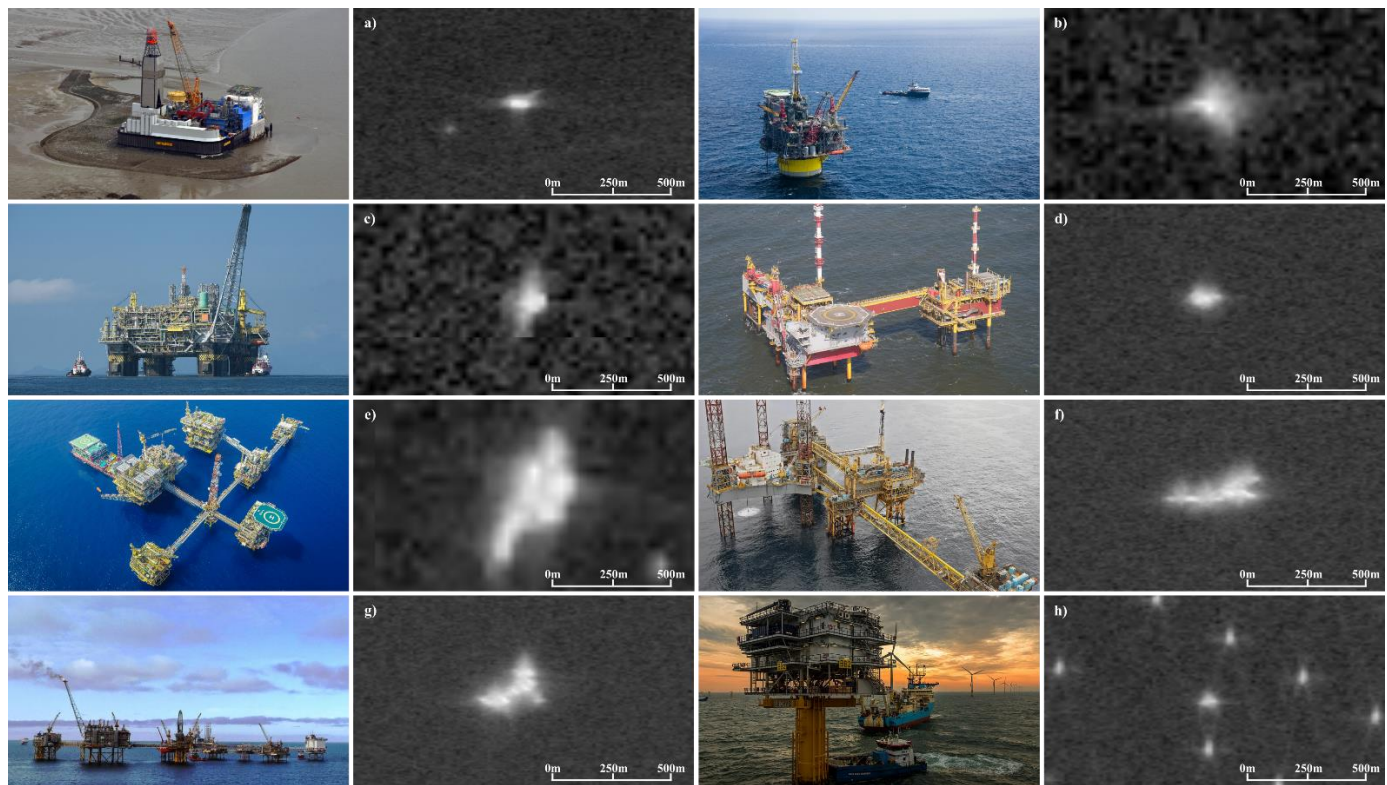

**Figure 3.** Complexity of platform occurrences. Sentinel-1 radar and aerial image of a diverse range of single platforms (a, b), more complex platform clusters (d-g), and a wind farm with turbines and substation (h). Aerial imagery retrieved from commons.wikimedia.org, credit to [41–44].

## 2.3. Ground truth data

For model training and testing, a comprehensive and consistent ground truth (GT) dataset was established. As explained in the previous chapter, some of the platforms are complex structures consisting of several individual platforms (referring hereinafter as platform clusters) (Figure 3d-g). Platform clusters comprise all occurrences consisting of more than one individual platform; in other words, contiguous platform complexes (Figure 3d-g). Multiple closely spaced backscatter signals from individual platforms result in stronger and blended noise in the S1 radar image, which manifests itself as a coherent complex backscatter signal. Individual platforms can hardly be derived from this, which is why we have labelled each complex as a single object. The goal of this study was to identify platforms in general. We did not label these objects as platforms, but rather as platform clusters to test whether we could optimise our model performance by treating single platforms and platform clusters separately during training and then combining them into a unified platform class for evaluation. We therefore defined three classes: *single platforms*, *platform clusters*, and *wind turbines* (off-targets).

Label creation was performed manually, drawing segmentation masks around radar backscatter signatures of the three classes. For training, bounding boxes (bboxes) were derived from these vector masks. The primary reference was Sentinel-1 GRD imagery from 2023Q4, supported by high-resolution optical data from DigitalGlobe (Google Earth) and auxiliary datasets such as vector layers and web-based infrastructure databases (Table 2).

**Table 2**. Auxiliary data used for the labelling process of the ground truth dataset.

| Dataset | Year | Coverage | Source |
|---|---|---|---|
| S1 GRD SAR imagery | 2023 | globally | [36] |
| DigitalGlobe imagery | - | globally | |
| Global Energy Monitor | 2025 | globally | [45] |
| OpenSeaMap | 2025 | globally | [46] |
| 4C Offshore | 2025 | globally | [47] |
| North Sea Energy | 2022 | North Sea | [48] |
| OSPAR | 2023 | North Sea | [49] |
| BOEM | 2024 | Gulf of Mexico | [50] |

Cross-validation among multiple data sources ensured positional accuracy and minimised labelling errors, providing a reliable foundation for consistent ground truth generation. The final GT dataset for all seven areas comprised more than 14,000 polygons (41% *single platform*, 4% *platform cluster*, 55% *wind turbine*) (Table 3).

To avoid data leakage and ensure independent evaluation, training, and test datasets of pairs of image chips and labels (referring hereinafter as image-label pairs) were kept strictly separate. All GT image-label pairs from the SCS, CS, GoG, and CoB were used for model training, while pairs from the NS, GoM, and PG served exclusively for testing.

The image-label pairs used for model training were randomly divided into 90% training and 10% validation subsets to track model performance during training. We ensured that image chips originating from the same image tile were not split between sets. The entire set of 8,909 labels from NS, GoM, and PG constituted the fixed test dataset, which remained constant across all experiments to ensure comparability.

The entire training set comprised 5,521 labels, with only 271 labels belonging to the *platform cluster* class (Table 3). This highlights the significant underrepresentation of this class, which, if not balanced, would disappear in the training signal due to the dominance of single platforms and wind turbines. To improve training stability and model generalisation, we aimed for an approximately balanced distribution among the three classes. Since obtaining additional labels was not feasible due to the limited availability of real-world samples, two potential strategies remained: expanding the dataset with *platform clusters* from new regions or generating synthetic data to increase the number of image-label pairs.

**Table 3**. Ground truth data, labels per region.

| Study regions | *single platform* | *platform cluster* | *wind turbines* | sum ∑ |
|---|---|---|---|---|
| NS | 418 | 80 | 4,949 | 5,447 |
| PG | 1,550 | 163 | | 1,713 |
| GoM | 1,562 | 187 | | 1,749 |
| SCS | 1,054 | 98 | 2,920 | 4,072 |
| CS | 311 | 81 | | 392 |
| GoG | 932 | 89 | | 1,021 |
| CoB | 33 | 3 | | 36 |

## 2.4. Synthetic data generation

We wanted to generate synthetic data and integrate it into training to optimise and improve our training data through better generalisation. The effects of this are a key focus of this study. With the synthetic data, we wanted to increase the number of image-label pairs, especially for the underrepresented class of *platform clusters*. Instead of oversampling the class by duplication of samples together with data augmentation (such as clipping, rotation, scaling, and mosaicking), we expected significantly better detection performance by generating images that are completely new to the model. Therefore, the goal was to create synthetic data that closely resembled real-world, hand-labelled S1 samples in appearance and diversity. For that task, we used the SyntEO framework [51], a comprehensive toolkit for creating custom synthetic EO data.

SyntEO composes a remote sensing scene from 2D discrete objects and background entities. Sensor-specific textures are then applied to simulate radar signatures, while the geometric composition provides bbox annotations. We extended the original SyntEO toolset to include routines for generating platform cluster geometries, in addition to single platforms and wind turbines. Two aspects were crucial in the entire process of generating synthetic data: full automation and randomisation to produce thousands of diverse image-label pairs simultaneously, and minimising the risk of the model learning artificial or repetitive patterns. To ensure natural variability, the appearance and configuration of each synthetic object differed across scenes. The overall workflow is summarised in Figure 4, which illustrates the main processing stages adopted in this study. For a detailed technical description of SyntEO, refer to Hoeser and Kuenzer [51].

(1) Input data consisted of randomly selected S1 scenes in WGS84 projection (EPSG:4326) from around the world. We defined the image extent in pixels and ensured a representative mix of open-sea, coastal, and land scenes. Images containing interfering objects, such as existing platforms, wind turbines, aquaculture, or lighthouses, were removed using a thresholding process to obtain 'empty' marine area entities suitable for synthetic insertion.
(2) Images were processed with SyntEO to derive coherent 2D entities for sea, coast, and land. The land entity was determined from the OSM Land Polygon Vector dataset [52]. A 1 km buffer from land defined the coast entity, and the remaining area defined the sea entity. Within the sea entity, we created a grid of potential object anchor points (candidate entities). These points were randomised per scene, creating a 'canvas' for placing synthetic objects.
(3) To realistically reproduce platform clusters, we first analysed real S1 radar signatures to identify common geometric features. Each cluster could be effectively described as a set of lines and points arranged along those lines. We generated random cluster geometries defined by meta parameter values such as the number and length of lines, connection angles, and point spacing.
(4) Each geometry was randomly rotated and placed in the scene at one of the candidate points of our grid.
(5) Radar backscatter was simulated by applying two-dimensional (2D) kernels to the generated geometries. Kernel properties, size, orientation, intensity, and spatial distribution were randomised for every point, ensuring diverse, realistic texture patterns. The textured geometries were then inserted into the prepared S1 scenes, resulting in complete synthetic images. Bbox annotations were automatically derived from object extent and position. Representative examples comparing synthetic and real-world scenes are shown in Figure 5.

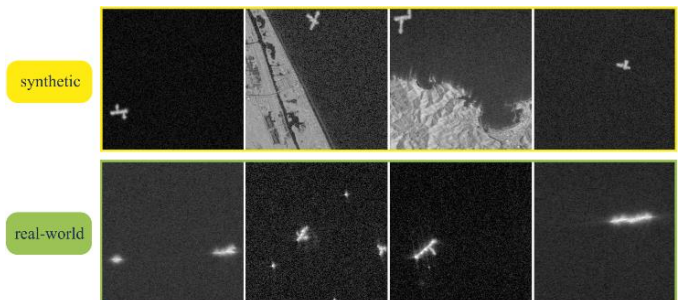

**Figure 5.** Comparison between synthetic scenes of platform cluster occurrences made by the SyntEO workflow and scenes from the real world.

## 2.5. Datasets

Five training datasets were compiled based on combinations of real and synthetic image-label pairs. The composition of these datasets is summarised in Table 4. The 'base' training dataset consisted of GT from SCS, CS, GoG, and CoB, with *single platforms* and *platform clusters* as a combined class, while the 'split' dataset split these two classes. To systematically assess the impact of synthetic training data, three additional datasets were created. In the cluster-enriched dataset, synthetic image-label-pairs were added to the underrepresented *platform cluster* class, resulting in a more balanced distribution across all three classes (2,330-2,920 pairs each). The fully balanced dataset was generated by augmenting all three classes with synthetic examples to reach 5,000 pairs per class. Finally, a synthetic-only dataset was produced to test model performance when trained exclusively on synthetic data.

For evaluation, a fixed test dataset containing real-world GT labels from the NS, PG, and GoM was used across all experiments to ensure consistency and comparability.

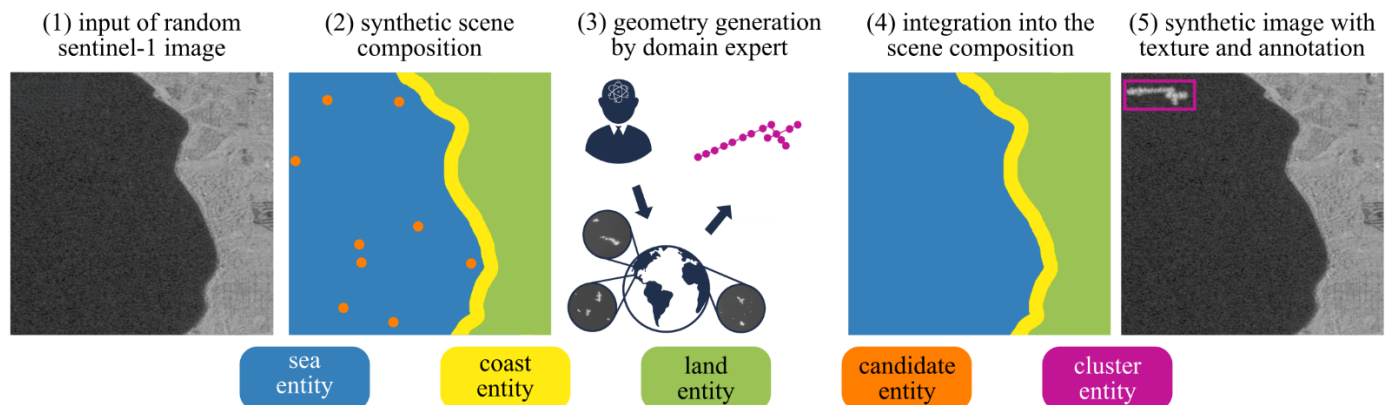

**Figure 4.** Simplified schematic overview of using the SyntEO framework to generate synthetic training data of platform clusters.

Table 4. Training and evaluation datasets with bounding box count, *marks synthetic training samples.

| Dataset | single platform | platform cluster | wind turbines | Sum ∑ |
|---|---|---|---|---|
| Base (SCS, CS, GoG, CoB) | 2,601 | | 2,920 | 5,521 |
| Split (SCS, CS, GoG, CoB) | 2,330 | 271 | 2,920 | 5,521 |
| Cluster-enriched | 2,330 | 271 + 2,206* | 2,920 | 7,727 |
| Fully balanced | 2,333 + 2,677* | 271 + 4,729* | 2,920 + 2,080* | 15,000 |
| Synthetic-only | 5,000* | 5,000* | 5,000* | 15,000 |
| Evaluation (NS, PG, GoM) | 3,530 | 430 | 4,949 | 8,909 |

## 2.6. Offshore platform detection pipeline

We employed version 10 of the YOLO OD algorithm family, released in May 2024 [53], to evaluate the effects of optimised training strategies for deep learning-based object detection in EO data and detecting offshore platforms in our test regions. YOLO is established and widely used in EO, offering a balance of accuracy, speed, and ease of use [54, 55]. The YOLOv10 model takes the S1 scenes as input, processes them through a CNN, and outputs the probabilities of the predefined classes: *single platform*, *platform cluster*, and *wind turbine*.

An overview of the detection workflow is shown in Figure 6, and the fundamentals are described in the following four stages.

(1) Model training: Various training scenarios were developed and tested under three distinct training scenarios: training solely with real data, training solely with synthetic data, and training with real and synthetic data combined (Table 4). The central research question was how synthetic data would influence model performance and geographic generalisation. We hypothesised that the targeted inclusion of synthetic samples would lead to a measurable improvement in accuracy and robustness. To prepare the trained models for Inference, the trained YOLOv10 artefacts were exported as a TorchScript model, packaged using the Torch-Model-Archiver [56], and uploaded as a deployable artefact.

(2) Inference: Model management and inference were carried out on the GCP. We utilised Vertex AI, GCP's machine-learning service, for model deployment and inference. Model endpoints were created on Vertex AI, and cloud CPUs were provisioned to perform inference over the preprocessed image chips. Importantly, all satellite data remained on the GCP; only prediction outputs were exported as text files containing detection metadata and geolocation. This design minimised data transfer and reduced cost by limiting billing to compute time only.

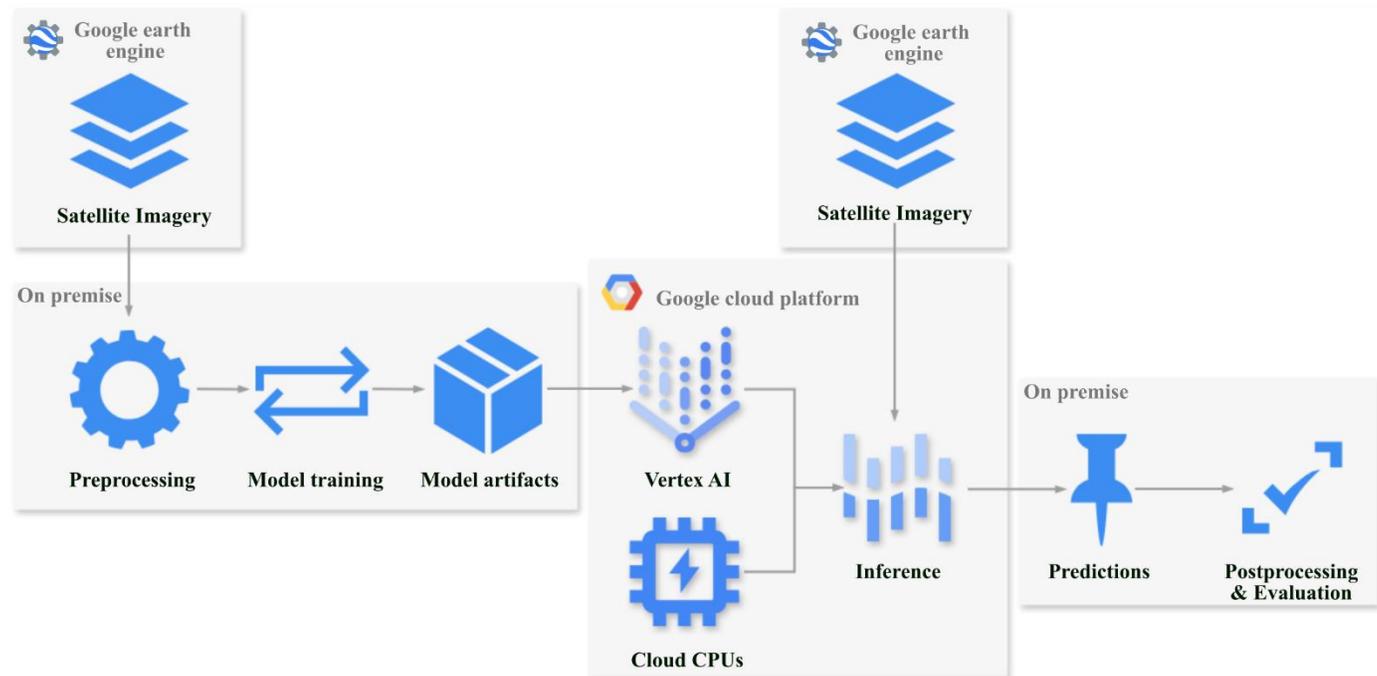

**Figure 6.** Simplified workflow of the deep learning-based offshore platform detection pipeline.

(3) Postprocessing and evaluation: After inference, the predictions from all image chips were merged into a single dataset. Bboxes were reprojected to EPSG:4326 (WGS84) to enable spatial operations, and each detection was assigned a unique identifier. Predictions below the defined confidence threshold were discarded. To suppress noise, detections where all pixels were below 150 (≈ -16.5 dB) were removed. This threshold was derived from the analysis of radar backscatter characteristics of verified platforms in the 8-bit S1 composites. To reduce duplicates caused by tile overlap, overlapping detections were grouped using an intersection-over-union (IoU) threshold of 0.2. For each group, the most reliable prediction was selected based on class consensus and confidence score. The cleaned detection set was exported as GeoJSON for further spatial analysis.

## 3. Results

To verify the effectiveness of the methods developed in this study, various training experiments were conducted and tested, resulting in a model that efficiently and accurately detects offshore platforms on S1 radar images and generalises reliably to previously unseen regions, thus being geographically transferable and scalable. First, the results of the training experiments and the influence of synthetic training data are described before finally the results of the best model are presented.

### 3.1. Training experiment results and synthetic data influence

For the experiments, we used pretrained YOLOv10n and s, and v11n model variants [53] that were finetuned with the respective training datasets. After 50 epochs, training was finished by an early stopping mechanism, tracking the validation performance. For testing, we compared model predictions against the GT of our fixed test dataset using a confidence threshold of 0.5, ensuring that only predictions with a probability of over 50% were considered. Further, predictions overlapping a GT object with a higher IoU value of 0.3 were counted as TPs, while lower overlaps were counted as FPs, and undetected GT objects as FNs. The results of all experiments are summarised in Table 5.

In the first experiment, we employed the smallest YOLOv10 variant (10n, 2.3 million parameters), trained on the base dataset. Despite its compact size, the model achieved strong performance, correctly detecting 3,537 of 3,960 platforms (F1 = 0.82). Using the split dataset next allowed training of *single platforms* and *platform clusters* instead of them being combined. The intermediate results (before merging the *single platforms* and *platform clusters* into a uniform class) showed that the model achieved an F1 score of 0.85 for *single platforms* and 0.42 for *platform clusters*. This highlights the model's difficulties in recognising *platform clusters*. However, these intermediate results still contain misclassifications between *single platforms* and *platform clusters*, which are eliminated in the final result by merging the classes. With an F1 score of 0.84, which is higher than in the first experiment, our assumption that the best recognition results for platforms can be achieved if the classes are first separated and only merged again after training and inference was confirmed.

**Table 5**. Evaluation of the training experiments on the evaluation dataset.

| Dataset | YOLO model | Class | GT | TP | FP | FN | Pr | Rc | F1 |
|---|---|---|---|---|---|---|---|---|---|
| Base | 10n *(2.3M params)* | single platform + platform cluster | 3,960 | 3,537 | 1,115 | 423 | 0.76 | **0.89** | 0.82 |
| Split | 10n *(2.3M params)* | single platform | 3,530 | 2,983 | 509 | 547 | 0.85 | 0.85 | 0.85 |
| | | platform cluster | 430 | 173 | 223 | 257 | 0.44 | 0.40 | 0.42 |
| | | platform | 3,960 | 3,319 | 569 | 641 | **0.85** | 0.84 | **0.84** |
| Split | 11n *(2.6M params)* | single platform | 3,530 | 3,011 | 552 | 519 | 0.85 | 0.85 | 0.85 |
| | | platform cluster | 430 | 274 | 479 | 156 | 0.36 | 0.64 | 0.46 |
| | | platform | 3,960 | 3,475 | 841 | 485 | 0.81 | **0.88** | 0.84 |
| Split | 10s *(7.2M params)* | single platform | 3,530 | 3,025 | 567 | 505 | 0.84 | 0.86 | 0.85 |
| | | platform cluster | 430 | 210 | 242 | 220 | 0.46 | 0.49 | 0.48 |
| | | platform | 3,960 | 3,387 | 657 | 573 | 0.84 | **0.86** | 0.85 |
| Cluster-enriched | 10s *(7.2M params)* | single platform | 3,530 | 3,036 | 419 | 494 | 0.88 | 0.86 | 0.87 |
| | | platform cluster | 430 | 240 | 63 | 190 | 0.79 | 0.56 | 0.65 |
| | | platform | 3,960 | 3,412 | 346 | 548 | **0.91** | 0.86 | 0.88 |
| Fully balanced | 10s *(7.2M params)* | single platform | 3,530 | 3,083 | 438 | 447 | 0.88 | 0.87 | 0.87 |
| | | platform cluster | 430 | 255 | 95 | 175 | 0.73 | 0.59 | 0.65 |
| | | platform | 3,960 | 3,523 | 348 | 437 | **0.91** | 0.89 | **0.90** |
| Synthetic-only | 10s *(7.2M params)* | single platform | 3,530 | 11 | 54 | 3,519 | 0.17 | 0.00 | 0.01 |
| | | platform cluster | 430 | 0 | 0 | 430 | 0.00 | 0.00 | 0.00 |
| | | platform | 3,960 | 13 | 52 | 3,947 | 0.20 | 0.00 | 0.01 |

We therefore retained this approach for the following experiments.

Using YOLOv11n (2.6 million parameters) in the next experiment did not improve the results, while switching to larger weights (v10s, 7.2 million parameters) slightly increased the F1 score to 0.85. Although even larger models (m, l, x) might have yielded marginal gains, we decided against them to ensure scalability when performing inference on GCP or other compute environments.

The subsequent experiments evaluated the influence of synthetic data on model performance. Since the *platform cluster* class was severely underrepresented and achieved F1 scores below 0.48 in the previous models, we enriched the training data with over 2,000 synthetic cluster samples to balance the class distribution. This adjustment is particularly evident in the intermediate results: for the *platform clusters*, precision improved significantly from 0.46 to 0.79 and the F1 score from 0.48 to 0.65, with 30 more TPs and 179 fewer FPs. This improvement carried over to the overall result for the unified *platform* class (Pr = 0.91, Rc = 0.86, and F1 = 0.88), the best result at this stage. To test whether a wider balance across all classes would further improve performance, the training data was expanded to 5,000 samples per class with synthetic samples. This approach improved recall in the intermediate results of the *single platforms* and *platform clusters* while maintaining high precision, leading to our best overall performance of the unified *platform* class (Pr = 0.91, Rc = 0.89, F1 = 0.90). Finally, to determine the upper limit of the usability of the synthetic data we generated, the model was trained with 5,000 synthetic samples per class exclusively. With this approach, we were unable to generalise the model to real images. The model achieved precision and recall close to zero. The model was apparently unable to learn or derive the features of the classes from synthetic data alone, which it needed in order to subsequently recognize the objects in the real data. We can assume that the model only learned synthetic specific features that could not be transferred to real scenes. Accordingly, we suspect a potential bottleneck in the data generated with SyntEO up to this point. Improving the designed textures could lead to significant detection performance [32, 51] and further increase performance when integrated with real-world data.

The experiments demonstrate both the effectiveness and limitations of synthetic data in training deep learning models for EO applications. Synthetic data significantly improved performance when combined with real-world data, particularly through class balancing. However, real-world examples remained indispensable.

### 3.2. Detection results and geographic transferability

The 'fully balanced" model achieved the best overall performance and was used to extract offshore platforms in the three unseen test regions of the NS, GoM, and PG. Figure 8 provides an overview of the detections of offshore platforms in these regions for 2023Q4 and shows the spatial distribution of this offshore infrastructure for these high-energy-producing regions. In total, the model detected 3,529 offshore platforms, corresponding to a recall of 0.89 and an F1 score of 0.90. The detection performance varied slightly from region to region: 411 platforms were detected in the NS (Rc = 0.83), 1,519 in the GoM (Rc = 0.87), and 1,593 in the PG (Rc = 0.93).

Geographically, the detection maps show that the model was able to reliably generalise to previously unseen regions, proving its robust transferability beyond the training areas (CS, SCS, GoG, and CoB). The dense arrangement of platforms in the central NS, the northern GoM, and the agglomerations in the oil fields in the PG is captured with high spatial accuracy, and even smaller installations near the coast are reliably detected.

The model showed high sensitivity to different platform types, from single installations to complex platform clusters. These results demonstrate the model's ability to adapt to varying regions with different environmental conditions, detection geometries, and structural complexities, confirming its potential for scalable global mapping of offshore infrastructure. Notably, it was possible to detect the off-target class of *wind turbines* with an F1 score of 0.97, which allowed them to be separated from platforms effectively.

Distinguishing substations in offshore wind farms from turbines is a crucial step of particular importance for refining the detection of offshore wind farms [30, 31, 57].

Figure 7 shows the corresponding confusion matrix of the final model. It shows high accuracy for both the *platform* class and the *wind turbine* off-target class (89% for *platforms*, 96.6% for *wind turbines*). Misclassifications were low and are discussed in Section 4.

| ground truth | platform | windturbine | background FN |
|---|---|---|---|
| platform | 3,523 / 3,960  89.0% | 0.4%  15 | 10.7%  422 |
| windturbine | 0.7%  36 | 4,783 / 4,949  96.6% | 2.6%  130 |
| background FP | 312 | 80 | |
| | platform | windturbine | background FN |
| | | predictions | |

**Figure 7.** Confusion matrix of the postprocessed detection results of the final model.

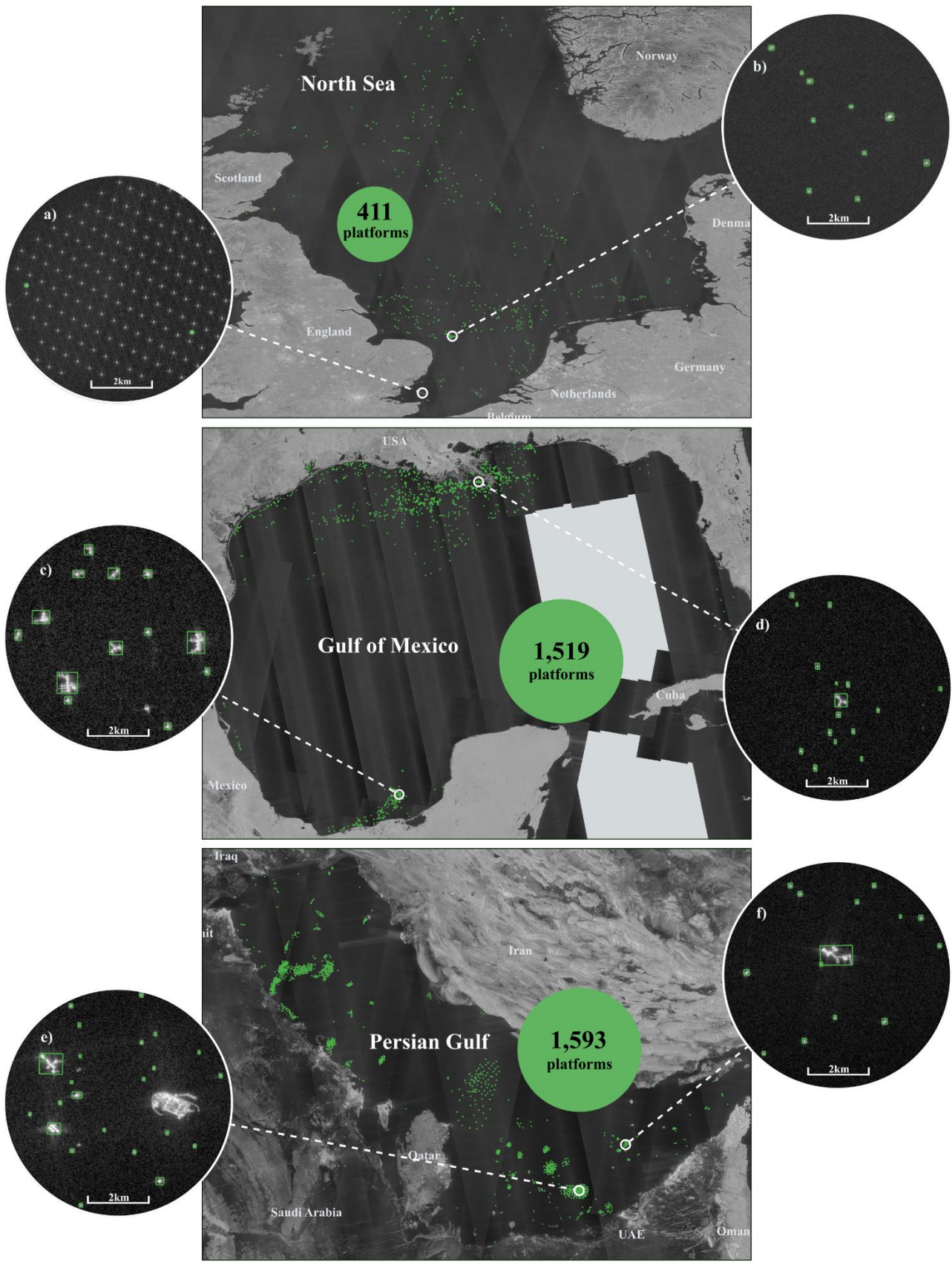

**Figure 8.** Offshore platform detection results in the three test regions North Sea, the Gulf of Mexico, and the Persian Gulf. The enlarged images show various challenging areas that are often crowded with different types of platforms (b-f). Wind turbines could be effectively separated from offshore platforms such as substations in a wind farm (a).

Overall, the results confirm that YOLO-based object detection on S1 radar images provides a reliable and transferable framework for the automatic identification of offshore platforms in various marine environments.

## 4. Discussion

In all three test regions, the North Sea, Gulf of Mexico, and Persian Gulf, 3,523 of 3,960 platforms (89%) were correctly identified, with 422 unidentified (FN) and 312 platforms predicted where none exist (FP). The model performed better on the off-target class of *wind turbines*: 4,783 of 4,949 turbines (96.6%) were correctly identified, with only 51 cross-class misclassifications. These results underscore that while the model generalises reliably across all classes, the detection of platforms remains the more complex task.

An important factor affecting FNs was the trade-offs made with the chosen confidence and IoU thresholds. Strict thresholds reduced the number of FPs, but unavoidably excluded some TPs with lower confidence. Through iterative tuning, a balance was found that minimised overall error. Additional FNs were associated with the detection of very small platforms, particularly in the Gulf of Mexico (Figure 9d), where numerous structures exist that are smaller than 100 m². At a spatial resolution of 10 m, such targets approach or fall below the effective pixel scale of S1, making their reliable detection difficult.

Misclassifications were mainly associated with objects that generate radar backscatter similar to that of platforms. These include, for example, buoys, lighthouses, and coastal reefs (Figure 9a-b). Including additional off-target classes in the training dataset could help to avoid such confusion for the model. However, the limited availability of real-world reference data would likely require supplementation with synthetic samples.

In a few cases in the NS, the model also incorrectly classified wind turbines under construction as platforms (Figure 9c). These unfinished turbines, which consist only of the foundation, are very similar to offshore platforms in both geometry and radar response.

Despite these challenges, the model performed particularly well in areas with dense and complex offshore infrastructure, such as the AKAL oil field in the southern GoM or the Upper Zakum field off the coast of the United Arab Emirates, where there are several platform clusters and artificial islands (Figure 8c, e). The accurate detection of such clustered targets demonstrates the model's ability to represent structural diversity and spatial context, capturing both individual installations and large platform complexes.

Another aspect concerns the incorporation of synthetic training data, which has proven to be crucial for the overall performance of the model. While synthetic samples significantly improved performance, especially for underrepresented classes, experiments with purely synthetic training data revealed distinct limitations. As stated in the previous chapter, we suspect a potential bottleneck in the data generated with SyntEO, which could be resolved by further improving the designed textures, e.g., through more nuanced simulation of backscattering, texture generation, or noise modelling. Further, generative approaches where the texture representation is learned instead of hand-crafted could be promising for scaling high-quality synthetic datasets with less manual effort.

Overall, the results show that YOLO-based object detection on S1 radar images is both robust and transferable to different marine environments, but sensitive to very small structures and class ambiguity. Addressing these remaining challenges through targeted dataset refinement and advanced synthetic data generation will be key to achieving consistent mapping of offshore infrastructure on a global scale.

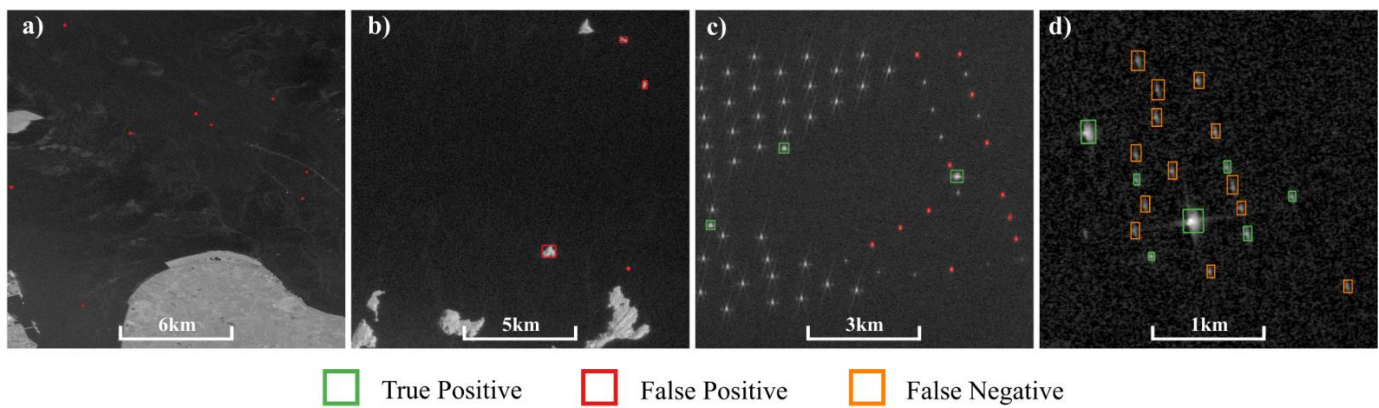

**Figure 9.** Demonstration of exemplary misclassifications of the model in the test regions. Misinterpretations of buoys and a lighthouse near Bremerhaven, Germany in the North Sea (a)), reefs at the northern coast of Oman in the Persian Gulf (b)), wind turbine foundations at the Gode wind farm north of the Norderney island, Germany in the North Sea (c)), and undetected small platforms to the southeast of New Orleans in the northern Gulf of Mexico (d)).

# 5. Conclusion

In the fourth quarter of 2023, we detected a total of 3,529 offshore platforms in three of the most important regions of the global offshore energy infrastructure: 411 in the North Sea, 1,519 in the Gulf of Mexico, and 1,593 in the Persian Gulf. The objects were detected by applying deep learning to ESA's S1 radar data. In our study, we demonstrated an approach that accomplishes the optimisation of the training process of the deep learning-based object detector by integrating synthetic images with real EO data. Both individual and complex groups of platforms are accurately detected, and substations are reliably differentiated from turbines in offshore wind farms. The results demonstrate the high robustness, generalizability, and transferability of the approach to previously unseen areas.

The presented method contributes to addressing the current lack of up-to-date and complete data on offshore infrastructure, which is crucial for resource management, environmental protection, and marine surveillance. It provides a first step toward an objective and scalable tool that can monitor changes in the inventory and development of offshore platforms across broad spatial and temporal scales. Overall, this approach marks an important step forward and offers new insights into the transition to global, fully automated identification and monitoring of maritime energy infrastructure. It creates a solid foundation for improving government and scientific databases, supports more sustainable resource management, and contributes to transparency and traceability of changes in the world's oceans. Future work will focus on further optimising the proposed method and expanding its applicability. A key priority will be to transfer the model to other geographical regions or on a global scale, as well as to different temporal scales, to systematically record, analyse, and evaluate changes in maritime energy infrastructure.

## Acknowledgment

The authors gratefully acknowledge ESA's Copernicus program for providing free access to the Sentinel-1 data and the Google Earth Engine platform for preprocessing and making the data accessible.

## Disclosure statement

The authors report no conflict of interest.

## Funding

This research received no external funding.

## Author contribution

**Robin Spanier**: Conceptualisation, methodology, code development, data curation, validation, visualisation, original manuscript writing. **Thorsten Hoeser**: Methodology, code development, manuscript reviewing. **Claudia Kuenzer**: Supervision, manuscript reviewing.

## References

[1] Y. Liu, C. Hu, Y. Dong, B. Xu, W. Zhan, and C. Sun, "Geometric accuracy of remote sensing images over oceans: The use of global offshore platforms," *Remote Sensing of Environment*, vol. 222, pp. 244–266, 2019, doi: 10.1016/j.rse.2019.01.002.

[2] A. B. Bugnot *et al.*, "Current and projected global extent of marine built structures," *Nat Sustain*, vol. 4, no. 1, pp. 33–41, 2021, doi: 10.1038/s41893-020-00595-1.

[3] X. Zhang, M. Qiu, S. Tao, X. Ge, and M. Wang, "OPDNet: An Offshore Platform Detection Network Based on Bitemporal Bimodal Remote-Sensing Images and a Pseudo-Siamese Structure," *IEEE J. Sel. Top. Appl. Earth Observations Remote Sensing*, vol. 18, pp. 6409–6421, 2025, doi: 10.1109/JSTARS.2025.3539674.

[4] K. Sadeghi, "An overview of design, analysis, construction and installation of offshore petroleum platforms suitable for Cyprus oil/gas fields," *GAU J. Soc. Appl. Sci*, vol. 2, no. 4, pp. 1–16, 2007.

[5] OECD, Organisation for Economic Co-operation and Development, *The Ocean Economy to 2050*.

[6] R. W. K. Potter and B. C. Pearson, "Assessing the global ocean science community: understanding international collaboration, concerns and the current state of ocean basin research," *npj Ocean Sustain*, vol. 2, no. 1, 2023, doi: 10.1038/s44183-023-00020-y.

[7] J. Wang, M. Li, Y. Liu, H. Zhang, W. Zou, and L. Cheng, "Safety assessment of shipping routes in the South China Sea based on the fuzzy analytic hierarchy process," *Safety Science*, vol. 62, pp. 46–57, 2014, doi: 10.1016/j.ssci.2013.08.002.

[8] P. Ma, M. Macdonald, S. Rouse, and J. Ren, "Automatic Geolocation and Measuring of Offshore Energy Infrastructure With Multimodal Satellite Data," *IEEE J. Oceanic Eng.*, vol. 49, no. 1, pp. 66–79, 2024, doi: 10.1109/JOE.2023.3319741.

[9] P. E. Posen, K. Hyder, M. Teixeira Alves, N. G. Taylor, and C. P. Lynam, "Evaluating differences in marine spatial data resolution and robustness: A North Sea case study," *Ocean & Coastal Management*, vol. 192, p. 105206, 2020, doi: 10.1016/j.ocecoaman.2020.105206.

[10] B. J. Williamson, P. Blondel, E. Armstrong, P. S. Bell, C. Hall, and J. J. Waggitt, "A Self-Contained Subsea Platform for Acoustic Monitoring of the Environment Around Marine Renewable Energy Devices–Field Deployments at Wave and Tidal Energy Sites in Orkney, Scotland," *IEEE J. Oceanic Eng.*, vol. 41, no. 1, pp. 67–81, 2016, doi: 10.1109/JOE.2015.2410851.

[11] OECD, Organisation for Economic Co-operation and Development, "OECD Science, Technology and Industry Working Papers," 2021.

[12] D. March, K. Metcalfe, J. Tintoré, and B. J. Godley, "Tracking the global reduction of marine traffic during the COVID-19 pandemic," *Nature communications*, vol. 12, no. 1, p. 2415, 2021, doi: 10.1038/s41467-021-22423-6.

[13] European Commission; Directorate-General for Environment, *Environmental impact assessment of*

*projects – Rulings of the Court of Justice of the European union*: Publications Office of the European Union, 2022.

[14] J. Virdin *et al.,* "The Ocean 100: Transnational corporations in the ocean economy," *Science advances*, vol. 7, no. 3, 2021, doi: 10.1126/sciadv.abc8041.

[15] B. A. Wong, C. Thomas, and P. Halpin, "Automating offshore infrastructure extractions using synthetic aperture radar & Google Earth Engine," *Remote Sensing of Environment*, vol. 233, p. 111412, 2019, doi: 10.1016/j.rse.2019.111412.

[16] C. Sun, Y. Liu, S. Zhao, and S. Jin, "Estimating offshore oil production using DMSP-OLS annual composites," *ISPRS Journal of Photogrammetry and Remote Sensing*, vol. 165, pp. 152–171, 2020, doi: 10.1016/j.isprsjprs.2020.05.019.

[17] Y. Liu, C. Sun, Y. Yang, M. Zhou, W. Zhan, and W. Cheng, "Automatic extraction of offshore platforms using time-series Landsat-8 Operational Land Imager data," *Remote Sensing of Environment*, vol. 175, pp. 73–91, 2016, doi: 10.1016/j.rse.2015.12.047.

[18] T. Hoeser and C. Kuenzer, "Global dynamics of the offshore wind energy sector monitored with Sentinel-1: Turbine count, installed capacity and site specifications," *International Journal of Applied Earth Observation and Geoinformation*, vol. 112, p. 102957, 2022, doi: 10.1016/j.jag.2022.102957.

[19] C. A. Baumhoer, A. J. Dietz, K. Heidler, and C. Kuenzer, "IceLines - A new data set of Antarctic ice shelf front positions," *Scientific data*, vol. 10, no. 1, p. 138, 2023, doi: 10.1038/s41597-023-02045-x.

[20] T. Esch *et al.,* "World Settlement Footprint 3D - A first three-dimensional survey of the global building stock," *Remote Sensing of Environment*, vol. 270, p. 112877, 2022, doi: 10.1016/j.rse.2021.112877.

[21] United Nations Conference on Trade and Development (UNCTAD), *Exploring space technologies for sustainable development.* [Online]. Available: https://unctad.org/publication/exploring-space-technologies-sustainable-development

[22] J. Yang *et al.,* "The role of satellite remote sensing in climate change studies," *Nature Clim Change*, vol. 3, no. 10, pp. 875–883, 2013, doi: 10.1038/nclimate1908.

[23] M. Amani *et al.,* "Google Earth Engine Cloud Computing Platform for Remote Sensing Big Data Applications: A Comprehensive Review," *IEEE J. Sel. Top. Appl. Earth Observations Remote Sensing*, vol. 13, pp. 5326–5350, 2020, doi: 10.1109/JSTARS.2020.3021052.

[24] N. Gorelick, M. Hancher, M. Dixon, S. Ilyushchenko, D. Thau, and R. Moore, "Google Earth Engine: Planetary-scale geospatial analysis for everyone," *Remote Sensing of Environment*, vol. 202, pp. 18–27, 2017, doi: 10.1016/j.rse.2017.06.031.

[25] R. Spanier and C. Kuenzer, "Marine Infrastructure Detection with Satellite Data—A Review," *Remote Sensing*, vol. 16, no. 10, p. 1675, 2024, doi: 10.3390/rs16101675.

[26] S. Casadio, O. Arino, and A. Minchella, "Use of ATSR and SAR measurements for the monitoring and characterisation of night-time gas flaring from off-shore platforms: The North Sea test case," *Remote Sensing of Environment*, vol. 123, pp. 175–186, 2012, doi: 10.1016/j.rse.2012.03.021.

[27] L. Cheng, K. Yang, L. Tong, Y. Liu, and M. Li, "Invariant triangle-based stationary oil platform detection from multitemporal synthetic aperture radar data," *J. Appl. Remote Sens*, vol. 7, no. 1, p. 73537, 2013, doi: 10.1117/1.JRS.7.073537.

[28] A. Marino, D. Velotto, and F. Nunziata, "Offshore Metallic Platforms Observation Using Dual-Polarimetric TS-X/TD-X Satellite Imagery: A Case Study in the Gulf of Mexico," *IEEE J. Sel. Top. Appl. Earth Observations Remote Sensing*, vol. 10, no. 10, pp. 4376–4386, 2017, doi: 10.1109/JSTARS.2017.2718584.

[29] J. Zhang, Q. Wang, and F. Su, "Automatic Extraction of Offshore Platforms in Single SAR Images Based on a Dual-Step-Modified Model," *Sensors (Basel, Switzerland)*, vol. 19, no. 2, 2019, doi: 10.3390/s19020231.

[30] W. Xu *et al.,* "Proliferation of offshore wind farms in the North Sea and surrounding waters revealed by satellite image time series," *Renewable and Sustainable Energy Reviews*, vol. 133, p. 110167, 2020, doi: 10.1016/j.rser.2020.110167.

[31] T. Zhang, B. Tian, D. Sengupta, L. Zhang, and Y. Si, "Global offshore wind turbine dataset," *Scientific data*, vol. 8, no. 1, p. 191, 2021, doi: 10.1038/s41597-021-00982-z.

[32] T. Hoeser, S. Feuerstein, and C. Kuenzer, "DeepOWT: a global offshore wind turbine data set derived with deep learning from Sentinel-1 data," *Earth Syst. Sci. Data*, vol. 14, no. 9, pp. 4251–4270, 2022, doi: 10.5194/essd-14-4251-2022.

[33] F. S. Paolo *et al.,* "Satellite mapping reveals extensive industrial activity at sea," *Nature*, vol. 625, no. 7993, pp. 85–91, 2024, doi: 10.1038/s41586-023-06825-8.

[34] Y. Qiu, X.-M. Li, L. Yan, and Z. Chen, "Synergic sensing of light and heat emitted by offshore oil and gas platforms in the South China Sea," *International Journal of Digital Earth*, vol. 17, no. 1, 2024, doi: 10.1080/17538947.2024.2441932.

[35] Flanders Marine Institute, "IHO Sea Areas, version 3," 2018.

[36] E. S. A. ESA, *Sentinel-1*. [Online]. Available: https://sentiwiki.copernicus.eu/web/sentinel-1

[37] K. El-Darymli, P. McGuire, D. Power, and C. Moloney, "Target detection in synthetic aperture radar imagery: a state-of-the-art survey," *J. Appl. Remote Sens*, vol. 7, no. 1, p. 71598, 2013, doi: 10.1117/1.JRS.7.071598.

[38] GEE, Google Earth Engine and E. S. A. ESA, *Sentinel-1 SAR GRD Collection (COPERNICUS/S1\_GRD).* [Online]. Available: https://developers.google.com/earth-engine/datasets/catalog/COPERNICUS_S1_GRD

[39] ESA, European Space Agency, "Sentinel-2 Products Specification Document," Technical Report S2-PDGS-TAS-DI-PSD, 2021. [Online]. Available: https://sentinels.copernicus.eu/documents/247904/0/Sentinel-2-product-specifications-document-V14-9.pdf

[40] C. R. Jackson and J. R. Apel, "Synthetic aperture radar: marine user's manual," 2004.
[41] R. Roletschek, *11-09-04-mittelplate-by-RalfR-18.jpg*.
[42] J. Komen, *Oil platform north of Ameland, Holland (9429668332).jpg*.
[43] A. N. Raudøy, *Bligh Bank 2 offshore substation BE 2016.jpg*.
[44] D. Petrobras, *Oil platform P-51 (Brazil).jpg*. [Online]. Available: https://commons.wikimedia.org/w/index.php?curid=5621984
[45] Global Energy Monitor, *Global Oil and Gas Extraction Tracker*. [Online]. Available: https://globalenergymonitor.org/projects/global-oil-gas-extraction-tracker/tracker-map/
[46] O. OSM, *OpenSeaMap*. [Online]. Available: http://openseamap.org/
[47] 4C Offshore, *Open Global Offshore Map*. [Online]. Available: https://map.4coffshore.com/offshorewind/
[48] North Sea Energy, *Project Atlas - An interactive atlas of the North Sea*. [Online]. Available: https://northseaenergy.projectatlas.app/atlas/introduction-energy/visibility
[49] OSPAR, Oslo and Paris Conventions, *Inventory of Offshore Installations - 2023*. [Online]. Available: https://odims.ospar.org/en/maps/map-inventory-of-offshore-installations-2023/
[50] BOEM, Bureau of Ocean Energy Management, *Oil & Gas Mapping and Data: Oil & Gas Infrastructure*. [Online]. Available: https://www.boem.gov/oil-gas-energy/mapping-and-data
[51] T. Hoeser and C. Kuenzer, "SyntEO: Synthetic dataset generation for earth observation and deep learning – Demonstrated for offshore wind farm detection," *ISPRS Journal of Photogrammetry and Remote Sensing*, vol. 189, pp. 163–184, 2022, doi: 10.1016/j.isprsjprs.2022.04.029.
[52] O. OSM, *Land polygons*. [Online]. Available: https://osmdata.openstreetmap.de/data/land-polygons.html
[53] A. Wang *et al.,* "YOLOv10: Real-Time End-to-End Object Detection," 2024.
[54] T. Hoeser and C. Kuenzer, "Object Detection and Image Segmentation with Deep Learning on Earth Observation Data: A Review-Part I: Evolution and Recent Trends," *Remote Sensing*, vol. 12, no. 10, p. 1667, 2020, doi: 10.3390/rs12101667.
[55] T. Hoeser, F. Bachofer, and C. Kuenzer, "Object Detection and Image Segmentation with Deep Learning on Earth Observation Data: A Review—Part II: Applications," *Remote Sensing*, vol. 12, no. 18, p. 3053, 2020, doi: 10.3390/rs12183053.
[56] Meta AI, *Torch-Model-Archiver*: PyTorch Foundation, 2025. [Online]. Available: https://github.com/pytorch/serve/blob/master/model-archiver/README.md
[57] Z. Xu, H. Zhang, Y. Wang, X. Wang, S. Xue, and W. Liu, "Dynamic detection of offshore wind turbines by spatial machine learning from spaceborne synthetic aperture radar imagery," *Journal of King Saud University - Computer and Information Sciences*, vol. 34, no. 5, pp. 1674–1686, 2022, doi: 10.1016/j.jksuci.2022.02.027.